\documentclass[10pt,twocolumn,letterpaper]{article}

\usepackage{cvpr}              

\usepackage[dvipsnames]{xcolor}

\definecolor{cvprblue}{rgb}{0.21,0.49,0.74}
\usepackage[pagebackref,breaklinks,colorlinks,citecolor=cvprblue]{hyperref}

\def\paperID{*****} 
\def\confName{CVPR}
\def\confYear{2024}

\title{MapVision: CVPR 2024 Autonomous Grand Challenge Mapless Driving Tech Report}

\author{
Zhongyu Yang$^{1}$,
Mai Liu$^{2}$,
Jinluo Xie$^{1}$,
Yueming Zhang$^{1}$,
Chen Shen$^{1}$,
Wei Shao$^{1}$, \\
Jichao Jiao$^{2}$,
Tengfei Xing$^{1}$,
Runbo Hu$^{1}$,
Pengfei Xu$^{1}$\\
$^1$ Didi Chuxing,
$^2$ Beijing University of Posts and Telecommunications \\
{\tt\small jayshenchen@didiglobal.com,}
{\tt\small  liumai@bupt.edu.cn}
}

\begin{document}
\maketitle
\begin{abstract}
Autonomous driving without high-definition (HD) maps demands a higher level of active scene understanding. In this competition, the organizers provided the multi-perspective camera images and standard-definition (SD) maps to explore the boundaries of scene reasoning capabilities. We found that most existing algorithms construct Bird's Eye View (BEV) features from these multi-perspective images and use multi-task heads to delineate road centerlines, boundary lines, pedestrian crossings, and other areas. However, these algorithms perform poorly at the far end of roads and struggle when the primary subject in the image is occluded. Therefore, in this competition, we not only used multi-perspective images as input but also incorporated SD maps to address this issue. We employed map encoder pre-training to enhance the network's geometric encoding capabilities and utilized YOLOX to improve traffic element detection precision. Additionally, for area detection, we innovatively introduced LDTR and auxiliary tasks to achieve higher precision. As a result, our final OLUS score is 0.58.

\end{abstract}    
\section{Introduction}
\label{sec:intro}

Mapless driving offers significant advantages for autonomous vehicles that do not rely on high-definition (HD) maps. It can adapt more quickly to real-world road changes and reduces the cost associated with manual map annotation. Therefore, addressing the challenges in this area has important practical significance.

This task involves taking multi-perspective images and standard-definition (SD) maps as input, requiring not only the perception of lanes and traffic elements but also the topology relationships among lanes and between lanes and traffic elements. In this competition, we not only used an innovative area prediction head borrowed from LDTR~\cite{ldtr} and auxiliary tasks from mapTRv2~\cite{liao2023maptrv2}, but also integrated the SD map into the BEV feature map and resolved the issue of abstract SD map embedding learning by introducing a map encoder pre-training task. Lastly, we utilized YOLOX~\cite{ge2021yolox} to enhance traffic element detection capabilities. Our final metrics on the leaderboard were 0.58.

\section{Method}
\begin{figure*}[htb]
	\begin{center}
		\includegraphics[width=1.0\linewidth]{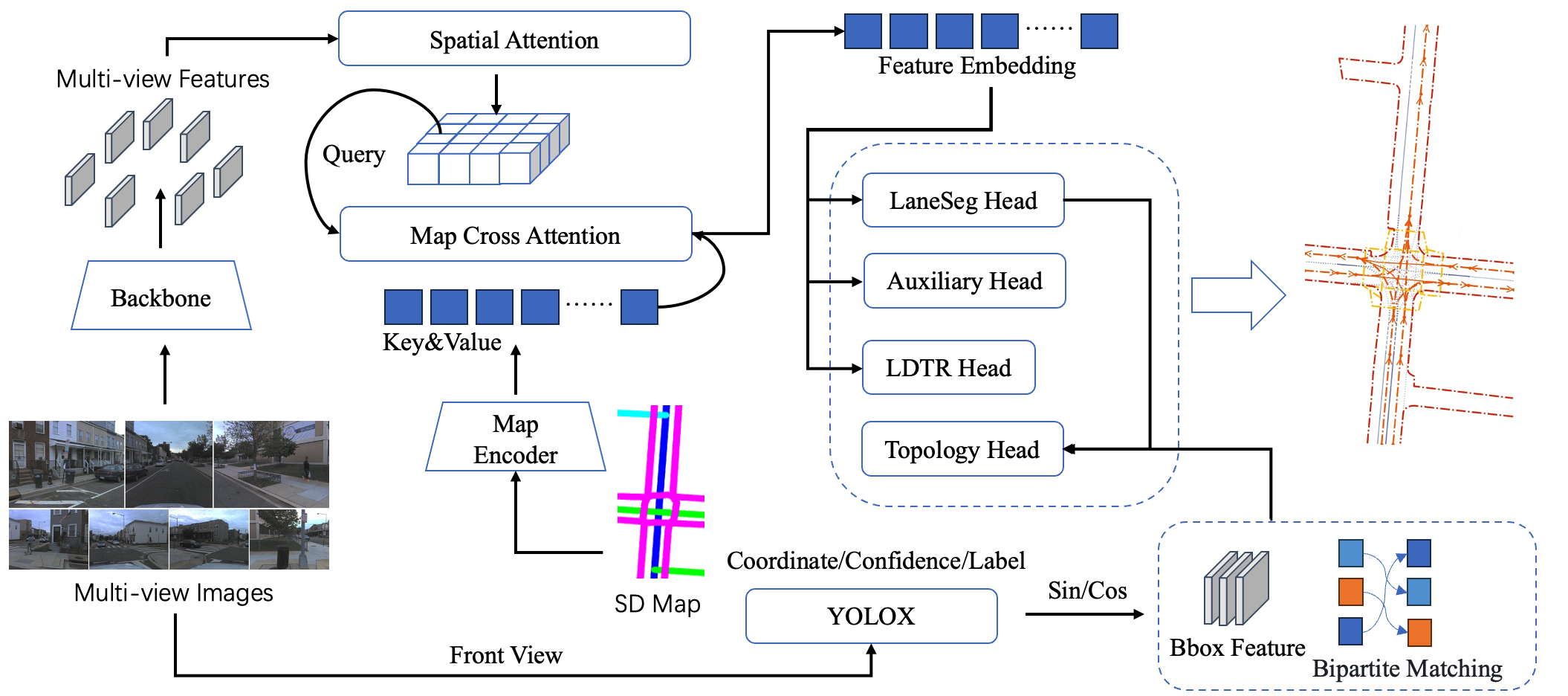}
	\end{center}
	\vspace{-0.5cm}
	\caption{The overall pipeline of our network}
	\vspace{-0.5cm}
	\label{fig:framework}
\end{figure*} 

\subsection{SD Map}
\begin{figure}[htb]
	\begin{center}
		\includegraphics[width=1.0\linewidth]{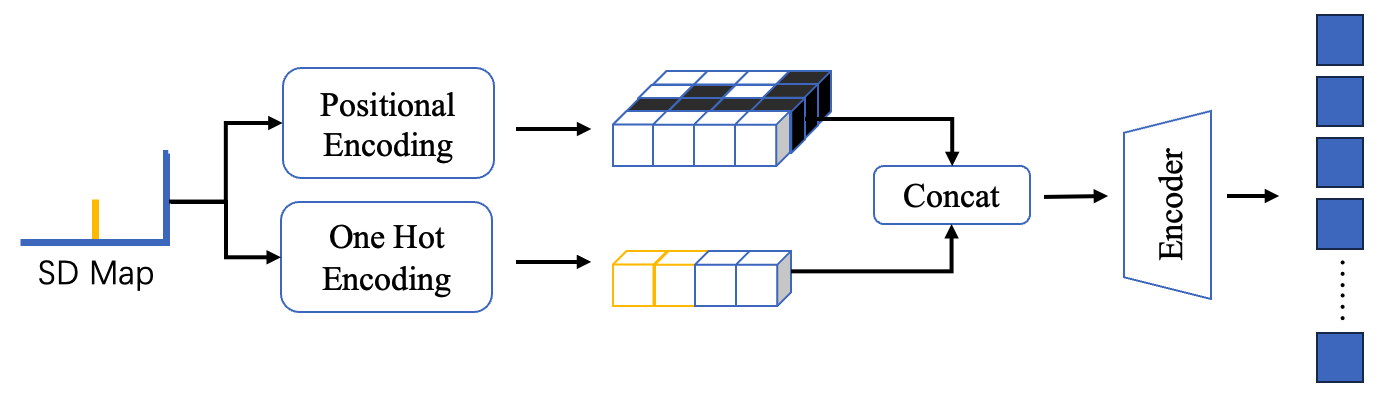}
	\end{center}
	\vspace{-0.4cm}
	\caption{Structure diagram of SD map encoding.}
	\vspace{-0.4cm}
	\label{fig:sdmap}
\end{figure}
SD maps are provided as supplementary elements to aid in understanding road topology from a BEV perspective and to offer map priors over longer distances.

\begin{table}[h] 
    \centering
    \small
    \resizebox{0.48\textwidth}{!}{
        \setlength\tabcolsep{4pt}
        \renewcommand\arraystretch{1.0}
        \begin{tabular}{lccccccc}
            \hline
             SD Map Sources & DET$_{l}$ & DET$_{a}$ & DET$_{t}$ & TOP$_{ll}$ & TOP$_{lt}$ & OLUS \\
            \hline
            \hline
              OpenStreetMap & 0.39 & 0.38 & 0.49 & 0.38 & 0.32 & 0.49\\
             OpenLaneV2 & 0.38 & 0.36 & 0.48 & 0.37 & 0.30 & 0.47\\
            \hline 
        \end{tabular}}
    \vspace{-6pt}
    \caption{Ablation experiments on different sources of SD map on OpenLaneV2 validation set.}
    \vspace{-2pt}
    \label{table:ablation on sdmap}
\end{table}

We encode maps from two SD map sources using the same network, and the final metrics are shown in \cref{table:ablation on sdmap}. So we utilized the open-source SD map from OpenStreetMap~\cite{haklay2008openstreetmap}. This map provides a large number of categories, a substantial amount of information, and comprehensive prior knowledge about roads. Each frame input into the model consists of a localized view of the SD map aligned with the vehicle's camera position. We maintain the structure of SMERF~\cite{luo2023augmenting}, as shown in \cref{fig:sdmap}, the map lines are encoded using sincos position encoding, while the category information is encoded with one-hot encoding. These features are concatenated and then processed through a Transformer encoder to learn map features.

After sincos position encoding of each line segment, the dimension of the graph vector becomes: 
\begin{equation}
  G_v = N_l * N_p * D
  \label{eq:graph_vector}
\end{equation}
where ${N}_{l}$ is the polylines in the view, ${N}_{p}$ is the number of default points, and ${D}$ is the dimension of the points, related to sincos position encoding and class embedding. After encoding with the Transformer encoder, the dimension of feature changes to 
\begin{equation}
  G_f = N_l * D_h
  \label{eq:graph_feature}
\end{equation}
where ${D}_{h}$ is the dimension of the hidden layer. The feature map ${G}_{f}$ is regarded as key and value, which is queried by the constructed BEV features.

SD map has structural prior information. To enhance the geometric structure encoding capability of the map encoder, we propose pre-training the map encoder. As shown in \cref{table:ablation on pretrain}, we compared the model with pre-trained weights loaded into the map encoder against the model without these weights.

In our experiments, we used an AutoEncoder approach where the feature embedding encoded with sincos position embedding was used as the ground truth. A lightweight decoder was used to predict the embedding with L2 loss as supervision. This approach significantly improved the line-related metrics. We also experimented with pre-training using the Masked AutoEncoder (MAE) method~\cite{he2022masked} but found no further improvement, possibly due to the default upsampling of 11 points per line.

\begin{table}[h] 
    \centering
    \small
    \resizebox{0.48\textwidth}{!}{
        \setlength\tabcolsep{4pt}
        \renewcommand\arraystretch{1.0}
        \begin{tabular}{lccccccc}
            \hline
            Method & DET$_{l}$ & DET$_{a}$ & DET$_{t}$ & TOP$_{ll}$ & TOP$_{lt}$ & OLUS \\
            \hline
            \hline
            Baseline  & 0.39 & 0.38 & 0.49 & 0.38 & 0.32 & 0.4913\\
            AutoEncoder Pre-train & 0.41 & 0.40 & 0.50 & 0.39 & 0.32 & 0.5008\\
            MAE Pre-train & 0.40 & 0.39 & 0.51 & 0.39 & 0.32 & 0.5001\\
            \hline 
        \end{tabular}}
    \vspace{-6pt}
    \caption{Ablation experiments on different pre-training methods on OpenLaneV2 validation set.}
    \vspace{-2pt}
    \label{table:ablation on pretrain}
\end{table}

\subsection{BEV Feature Constructor}

Based on BEVFormer~\cite{bevformer}, we project the perspective view (PV) features onto BEV and then fuse the image features with SD map features in the BEV domain. Additionally, we designed auxiliary foreground segmentation tasks for both PV and BEV to enhance feature extraction capabilities.

\subsection{Area Detection}

We utilized the training framework of mapTR~\cite{maptr} for area detection, working with BEV features that integrate surround-view and SD map information. However, we found that the approach in mapTR of using key points as queries weakened the overall integrity of instances. Therefore, drawing inspiration from LDTR~\cite{ldtr}, we employed an anchor-chain method to represent area instances and further optimized these instances holistically using MRDA and P2P IoU loss \& cost. The results indicate a significant performance improvement compared to mapTR.

\subsection{Lanesegment Detection}

For lanesegment detection, we primarily employed LaneSegNet~\cite{lanesegnet}, which conducts centerline and laneline detection on BEV features fused with surround-view and SD map information. Additionally, we incorporated P2P IoU loss and cost to enhance the overall optimization of lane segments.

\subsection{Traffic Element Detection}

\begin{table}[h] 
    \centering
    \small
    \resizebox{0.48\textwidth}{!}{
        \setlength\tabcolsep{36pt}
        \renewcommand\arraystretch{1.0}
        \begin{tabular}{lc}
            \hline
            Method &  DET$_{t}$ \\
            \hline
            \hline
            YOLOX  & 0.55 \\
            + Data Augmentation & 0.58 \\
            + PAFPN & 0.64 \\
            + Reweight & 0.66\\
            + Pseudo Label & 0.66\\
            + Data Resampling & 0.69 \\
            + TTA & 0.73 \\
            \hline 
        \end{tabular}}
    \vspace{-6pt}
    \caption{Ablation experiments of YOLOX on OpenLaneV2 validation set.}
    \vspace{-2pt}
    \label{table:yolox}
\end{table}

We utilized YOLOX~\cite{ge2021yolox} as our 2D object detection model, achieving higher quality bounding boxes. As shown in \cref{table:yolox}, throughout the training process, we experimented with various data augmentation techniques, data resampling, and pseudo-labeling methods, etc. Ultimately, we improved our metrics from a baseline of 0.55 to 0.73, marking a 33\% enhancement.

Firstly, we found that 40\% of the training data lacks the objects to be detected. Due to the sparsity of foreground samples, many images do not contain the target objects. To enable the model to learn more useful features, we employed two different data augmentation methods: mosaic and mixup.

Then, we incorporated the PAFPN~\cite{liu2018path} structure to fuse multi-scale features, which can address the sparsity issue of low-level features.

We found that the sample counts were unbalanced. Training samples for objects such as yellow lights, U-turns, and no U-turn signs were relatively scarce. To balance the number of positive examples for each detectable category, we performed data resampling and increased the classification weight for difficult samples.

We found that the length and width distribution of the bounding boxes is concentrated between 0 and 200, with an uneven size distribution. Smaller bounding boxes account for a larger proportion, while some larger-sized bounding boxes have ground truth values exceeding 200. To address this, we introduced Test Time Augmentation (TTA) and applied scale transformations to the inference images, ensuring good detection performance for objects of varying sizes.

Finally, to provide the detection model with more training data and enhance its generalization, We have annotated the validation set using pseudo-labels and included it in the training. This allowed the model to achieve better detection performance across the 13 classes of objects.

\subsection{Lane-Lane Topology}
By integrating features from the upstream backbone and the BEV features, and then passing them through the encoder in lanesegment, we extract the line features. We then perform bipartite matching with the ground truth line segments to obtain matched positive examples and unmatched negative examples. These examples are input into the topology head, where an adjacency matrix representing the matching relationship between lines is produced through an MLP.

\subsection{Lane-Traffic Topology}
To improve topology metrics, we fine-tuned the topology task head while keeping the rest of the network frozen. We imported the object detection results from YOLOX\cite{ge2021yolox} for each frame. By providing high-quality bounding boxes, we enhanced the quality of the detection features using sincos position encoding and the bipartite matching results compared with the original Deformable DETR~\cite{zhu2020deformable}. This, in turn, improved the prediction quality in the online inference of lane-traffic topology relationships, leading to a significant increase in the final precision.

\section{Experiments}
In this section, we provide some experimental details for reproducibility of the final results, which were evaluated on OpenLaneV2.

\subsection{Implementation Details}

We experimented with several backbone networks, including ResNet-50~\cite{he2016deep} and InternImage-L~\cite{wang2023internimage}. Our learning rate was 2e-4, the optimizer was AdamW~\cite{loshchilov2017decoupled}, and the total number of training epochs was 48 epochs. 

For traffic element detection, our image resolution is 1550*2480. 

For the SD map, we achieved better weight initialization by loading the pre-trained map encoder using OpenStreetMap~\cite{haklay2008openstreetmap}.

For topological reasoning, the learning rate was 4e-4. We froze the other parts of the network and only fine-tuned the heads for line and traffic element topology relationships.

\subsection{Model Performance}
We have made many improvements to the model, and now we list the effects of the methods that have proven to be effective in \cref{table:ablation on network}.

\begin{table}[h]
    \centering
    \small
    \resizebox{0.48\textwidth}{!}{
        \setlength\tabcolsep{4pt}
        \renewcommand\arraystretch{1.0}
        \begin{tabular}{lcccccc}
            \hline
            Method & DET$_{l}$ & DET$_{a}$ & DET$_{t}$ & TOP$_{ll}$ & TOP$_{lt}$ & OLUS \\
            \hline
            \hline
            ResNet-50  & 0.29 & 0.20 & 0.36 & 0.26 & 0.21 & 0.36\\
            + Map Encoder  & 0.35 & 0.24 & 0.40 & 0.29 & 0.21 & 0.40\\
            + InternImage-L & 0.38 & 0.25 & 0.49 & 0.32 & 0.30 & 0.45\\
            + Map Pretrain & 0.40 & 0.26 & 0.50 & 0.34 & 0.29 & 0.46\\
            + LDTR & 0.42 & 0.34 & 0.50 & 0.38 & 0.30 & 0.48\\
            + P2P IoU Loss & 0.43 & 0.37 & 0.51 & 0.41 & 0.31 & 0.50\\
            + Aux Head & 0.44 & 0.38 & 0.50 & 0.40 & 0.31 & 0.50\\
            \hline 
        \end{tabular}}
    \vspace{-6pt}
    \caption{Ablation experiments for different improvement methods on OpenLaneV2 validation set.}
    \vspace{-2pt}
    \label{table:ablation on network}
\end{table}

\noindent\textbf{Map Encoder.}
We used the SD map feature as the key and value, and the feature map formed from multi-perspective images as the query to perform cross-attention. We then found that with the prior of SD map, the quality of our HD map construction was greatly improved. Among them, DET$_{l}$ and DET$_{a}$ improved significantly, increasing by 19\% and 20\% respectively. The TOP$_{ll}$ metric increased by 13\%, DET$_{t}$ increased by 9\%, and TOP$_{lt}$ increased by 1.5\%. It can be seen that SD map can enhance the line features, and due to the inherent topological structure of the map, it also helps in constructing the final line-line topology graph.

\noindent\textbf{Large Backbone.}
We also tried a backbone with a larger number of parameters. We found that InternImage-L~\cite{wang2023internimage}, due to its DCNV3 convolutions, possesses the advantages of both CNN's inductive biases and multi-head attention. It can bring about stable and significant improvements across all metrics. Our final metric OLUS improved from 0.3960 to 0.4486, an increase of 13\%.

\noindent\textbf{Map Pretrain.}
To enhance the map encoder's understanding of maps, we pre-trained it. As a result, we found that the TOP$_{ll}$ metric improved the most, with an increase of 6\%, DET$_{l}$ increased by 3\%, DET$_{a}$ increased by 1.5\%, and DET$_{t}$ increased by 1.7\%. This is because during the pre-training task, the map encoder learned in advance the process of converting the position-encoded SD map into the feature map output by the encoder, leading to significant improvements in topological reasoning.

\noindent\textbf{LDTR Head.}
By introducing the LDTR~\cite{ldtr} head, our DET$_{a}$ improved the most, from 0.25 to 0.34, a 36\% increase. DET$_{l}$ increased by 4.5\%, TOP$_{ll}$ increased by 10\%, and TOP$_{lt}$ increased by 4.9\%.

\noindent\textbf{P2P IoU Loss.}
By introducing the P2P IoU loss, we incorporated the IoU metric between lines to supervise the matching degree of lines, enhancing the detection precision of lines. We found that DET$_{a}$ improved by 8\%, TOP$_{ll}$ increased by 7\%, DET$_{l}$ increased by 2.1\%, TOP$_{lt}$ increased by 4.6\%, and DET$_{t}$ increased by 1.9\%. It can be observed that the metrics related to lines have shown significant improvement.

\noindent\textbf{Aux Head.}
By introducing auxiliary task heads, although our overall metric did not increase on the validation set, there was a slight improvement on the test set.

\subsection{Model Ensemble}
As shown in \cref{table:final leaderboard}, we replaced the traffic element detection results in YOLOX with those from an existing model, and utilized the detection results from YOLOX to provide better features and binary matching results input through sincos position encoding. This allowed us to fine-tune the topology head, improving the TOP$_{lt}$ metric. Previous training results had indicated that when we increased DET$_{l}$ and DET$_{a}$, there was a decline in TOP$_{ll}$. We considered that this was due to insufficient model capacity. Therefore, we decoupled the training for the area head, which ultimately helped to improve DET$_{a}$.

\begin{table}[h] 
    \centering
    \small
    \resizebox{0.48\textwidth}{!}{
        \setlength\tabcolsep{4pt}
        \renewcommand\arraystretch{1.0}
        \begin{tabular}{lcccccc}
            \hline
            Method & DET$_{l}$ & DET$_{a}$ & DET$_{t}$ & TOP$_{ll}$ & TOP$_{lt}$ & OLUS \\
            \hline
            \hline
            Strong Baseline  & 0.44 & 0.38 & 0.50 & 0.40 & 0.31 & 0.50\\
            + YOLOX  & 0.44 & 0.38 & 0.73 & 0.40 & 0.46 & 0.57\\
            + Fintune Topology & 0.44 & 0.38 & 0.73 & 0.40 & 0.52 & 0.58\\
            + Decoupled Training & 0.44 & 0.42 & 0.73 & 0.40 & 0.52 & 0.59\\
            \hline
            Test Server & 0.39 & 0.40 & 0.80 & 0.38 & 0.48 & 0.58\\
            \hline 
        \end{tabular}}
    \vspace{-6pt}
    \caption{Ablation experiments for several model ensemble methods (First three rows: OpenLaneV2 validation set; Last row: OpenLaneV2 test set)}
    \vspace{-2pt}
    \label{table:final leaderboard}
\end{table}

\section{Conclusion}
In the paper, we propose a network that leverages the advantages of SD map and multi-view image inputs, combined with a mature object detector, and introduces SD map pre-training, the LDTR head, and auxiliary tasks to improve the final precision. As a result, as shown in the last row of \cref{table:final leaderboard}, the OLUS metric reached 0.58.

{
    \small
    \bibliographystyle{ieeenat_fullname}
    \bibliography{main}

\begin{thebibliography}{14}
\providecommand{\natexlab}[1]{#1}
\providecommand{\url}[1]{\texttt{#1}}
\expandafter\ifx\csname urlstyle\endcsname\relax
  \providecommand{\doi}[1]{doi: #1}\else
  \providecommand{\doi}{doi: \begingroup \urlstyle{rm}\Url}\fi

\bibitem[Ge et~al.(2021)Ge, Liu, Wang, Li, and Sun]{ge2021yolox}
Zheng Ge, Songtao Liu, Feng Wang, Zeming Li, and Jian Sun.
\newblock Yolox: Exceeding yolo series in 2021.
\newblock \emph{arXiv preprint arXiv:2107.08430}, 2021.

\bibitem[Haklay and Weber(2008)]{haklay2008openstreetmap}
Mordechai Haklay and Patrick Weber.
\newblock Openstreetmap: User-generated street maps.
\newblock \emph{IEEE Pervasive computing}, 7\penalty0 (4):\penalty0 12--18, 2008.

\bibitem[He et~al.(2016)He, Zhang, Ren, and Sun]{he2016deep}
Kaiming He, Xiangyu Zhang, Shaoqing Ren, and Jian Sun.
\newblock Deep residual learning for image recognition.
\newblock In \emph{Proceedings of the IEEE conference on computer vision and pattern recognition}, pages 770--778, 2016.

\bibitem[He et~al.(2022)He, Chen, Xie, Li, Doll{\'a}r, and Girshick]{he2022masked}
Kaiming He, Xinlei Chen, Saining Xie, Yanghao Li, Piotr Doll{\'a}r, and Ross Girshick.
\newblock Masked autoencoders are scalable vision learners.
\newblock In \emph{Proceedings of the IEEE/CVF conference on computer vision and pattern recognition}, pages 16000--16009, 2022.

\bibitem[Li et~al.(2023)Li, Jia, Wang, Chen, Jiang, Yan, and Li]{lanesegnet}
Tianyu Li, Peijin Jia, Bangjun Wang, Li Chen, Kun Jiang, Junchi Yan, and Hongyang Li.
\newblock Lanesegnet: Map learning with lane segment perception for autonomous driving.
\newblock \emph{arXiv preprint arXiv:2312.16108}, 2023.

\bibitem[Li et~al.(2022)Li, Wang, Li, Xie, Sima, Lu, Qiao, and Dai]{bevformer}
Zhiqi Li, Wenhai Wang, Hongyang Li, Enze Xie, Chonghao Sima, Tong Lu, Yu Qiao, and Jifeng Dai.
\newblock Bevformer: Learning bird’s-eye-view representation from multi-camera images via spatiotemporal transformers.
\newblock In \emph{European conference on computer vision}, pages 1--18. Springer, 2022.

\bibitem[Liao et~al.(2022)Liao, Chen, Wang, Cheng, Zhang, Liu, and Huang]{maptr}
Bencheng Liao, Shaoyu Chen, Xinggang Wang, Tianheng Cheng, Qian Zhang, Wenyu Liu, and Chang Huang.
\newblock Maptr: Structured modeling and learning for online vectorized hd map construction.
\newblock \emph{arXiv preprint arXiv:2208.14437}, 2022.

\bibitem[Liao et~al.(2023)Liao, Chen, Zhang, Jiang, Zhang, Liu, Huang, and Wang]{liao2023maptrv2}
Bencheng Liao, Shaoyu Chen, Yunchi Zhang, Bo Jiang, Qian Zhang, Wenyu Liu, Chang Huang, and Xinggang Wang.
\newblock Maptrv2: An end-to-end framework for online vectorized hd map construction.
\newblock \emph{arXiv preprint arXiv:2308.05736}, 2023.

\bibitem[Liu et~al.(2018)Liu, Qi, Qin, Shi, and Jia]{liu2018path}
Shu Liu, Lu Qi, Haifang Qin, Jianping Shi, and Jiaya Jia.
\newblock Path aggregation network for instance segmentation.
\newblock In \emph{Proceedings of the IEEE conference on computer vision and pattern recognition}, pages 8759--8768, 2018.

\bibitem[Loshchilov and Hutter(2017)]{loshchilov2017decoupled}
Ilya Loshchilov and Frank Hutter.
\newblock Decoupled weight decay regularization.
\newblock \emph{arXiv preprint arXiv:1711.05101}, 2017.

\bibitem[Luo et~al.(2023)Luo, Weng, Wang, Wu, Li, Weinberger, Wang, and Pavone]{luo2023augmenting}
Katie~Z Luo, Xinshuo Weng, Yan Wang, Shuang Wu, Jie Li, Kilian~Q Weinberger, Yue Wang, and Marco Pavone.
\newblock Augmenting lane perception and topology understanding with standard definition navigation maps.
\newblock \emph{arXiv preprint arXiv:2311.04079}, 2023.

\bibitem[Wang et~al.(2023)Wang, Dai, Chen, Huang, Li, Zhu, Hu, Lu, Lu, Li, et~al.]{wang2023internimage}
Wenhai Wang, Jifeng Dai, Zhe Chen, Zhenhang Huang, Zhiqi Li, Xizhou Zhu, Xiaowei Hu, Tong Lu, Lewei Lu, Hongsheng Li, et~al.
\newblock Internimage: Exploring large-scale vision foundation models with deformable convolutions.
\newblock In \emph{Proceedings of the IEEE/CVF Conference on Computer Vision and Pattern Recognition}, pages 14408--14419, 2023.

\bibitem[Yang et~al.(2024)Yang, Shen, Shao, Xing, Hu, Xu, Chai, and Xue]{ldtr}
Zhongyu Yang, Chen Shen, Wei Shao, Tengfei Xing, Runbo Hu, Pengfei Xu, Hua Chai, and Ruini Xue.
\newblock Ldtr: Transformer-based lane detection with anchor-chain representation.
\newblock \emph{arXiv preprint arXiv:2403.14354}, 2024.

\bibitem[Zhu et~al.(2020)Zhu, Su, Lu, Li, Wang, and Dai]{zhu2020deformable}
Xizhou Zhu, Weijie Su, Lewei Lu, Bin Li, Xiaogang Wang, and Jifeng Dai.
\newblock Deformable detr: Deformable transformers for end-to-end object detection.
\newblock \emph{arXiv preprint arXiv:2010.04159}, 2020.

\end{thebibliography}
}

\end{document}